# Efficient parametrization of multi-domain deep neural networks

Sylvestre-Alvise Rebuffi[1]  Hakan Bilen[2]  Andrea Vedaldi[1]

[1]Visual Geometry Group
University of Oxford
{srebuffi,vedaldi}@robots.ox.ac.uk

[2]School of Informatics
University of Edinburgh
hbilen@ed.ac.uk

## Abstract

*A practical limitation of deep neural networks is their high degree of specialization to a single task and visual domain. Recently, inspired by the successes of transfer learning, several authors have proposed to learn instead universal, fixed feature extractors that, used as the first stage of any deep network, work well for several tasks and domains simultaneously. Nevertheless, such universal features are still somewhat inferior to specialized networks.*

*To overcome this limitation, in this paper we propose to consider instead* universal parametric families of neural networks, *which still contain specialized problem-specific models, but differing only by a small number of parameters. We study different designs for such parametrizations, including series and parallel residual adapters, joint adapter compression, and parameter allocations, and empirically identify the ones that yield the highest compression. We show that, in order to maximize performance, it is necessary to adapt both shallow and deep layers of a deep network, but the required changes are very small. We also show that these universal parametrization are very effective for transfer learning, where they outperform traditional fine-tuning techniques.*

## 1. Introduction

As deep neural networks continue to dramatically improve results in almost all traditional problems in computer vision, the interest of the community has started to shift towards more ambitious goals. One of them is to supersede the common paradigm of addressing different image understanding problems independently, using ad-hoc solutions and learning different and largely incompatible models for each of them. Just like the human brain is capable of addressing a very large number of different image analysis tasks, so it should be possible to develop models that address well and efficiently a variety of different computer vision problems, with better efficiency and generalization

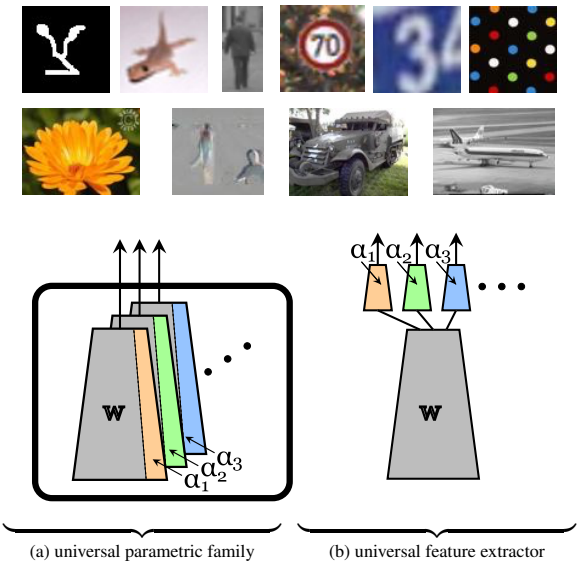

Figure 1: **Universal parametric network families.** We develop compact parametric families of neural networks (a) that can target very different visual domains, from ImageNet to stop signs and characters, while sharing the vast majority of their parameters **w**. Domain-specific parameters $\alpha_t$ are isolated in small modular adapters that can be attached to an existing network to steer it non-disruptively to different domains and enable efficient model storage, transfer, and exchange, as well as transfer learning. Parametric families are shown empirically to be much more powerful than sharing a fixed universal feature extractor as in (b).

than individual networks.

There are at least three aspects to this challenge. The first is to construct a *multi-task model* that can extract multiple types of information from an image, performing class/object detection and segmentation, boundary extraction, motion estimation, etc. [14]. The second is to develop a *multi-domain model* that can work well for many different visual domains, such as Internet images, scene text, medical images, satellite images, driving images, etc [3, 23]. The third is to develop an *extensible model* that can evolve over time, reusing previously acquired knowledge to learn to process new tasks and domains efficiently, while at the same time avoiding to forget previously-acquired abilities [18].

Concerned with the second and third problem, several authors before us have framed this as the problem of learning a single universal first-stage to be shared among different deep networks (fig. 1.b). The idea is that early layers should process low-level and hence widely-applicable visual information. However, such universal feature extractors do not work quite as well as learning problem-specific networks, either from scratch or using transfer learning.

In this paper, we propose an alternative perspective. Instead of seeking a single, fixed first stage, we want to develop compact parametrizations for multi-domain networks (fig. 1.a). Consider a deep network $\Phi(\mathbf{x}; \mathbf{w}, \alpha)$ applied to an image $\mathbf{x}$, for example for image classification. We partition the network parameters in a universal vector $\mathbf{w}$, which is fixed and shared among all domains, and a parameter vector $\alpha$, which is instead domain specific. We then seek architectures that: 1) can share the vast majority of their parameters, so that the size of $\alpha$ is a small fraction of the size of $\mathbf{w}$, and 2) can learn a new $\alpha$ for a new domain from a very small number of training examples. In other words, we would like to compress a family of domain-specific neural networks so that they can be exchanged and learned more efficiently.

While our method does not result in a single, universal neural network, as the parameters $\alpha$ are still domain-specific, finding architectures that afford a great degree of parameter sharing is an important step in this direction. There are also concrete practical benefits. First, universal families work better than the standard transfer learning approach of fine-tuning off-the-shelf models; hence, they may replace the latter strategy in numerous applications. Second, there are applications such as mobile devices that require running several different neural networks, which may incur a significant computational and energy overhead due simply to the need of swapping their parameters on a dedicated integrated circuit. One may face similar overheads when transmitting model parameters over a network, or storing them locally. Our approach makes storing, exchanging, and updating models much more efficient.

Related to our work, a few papers [23, 25] have proposed low-dimensional parametrizations of the filters in a neural network with good compression results. The paper of [23], in particular, proposed the idea of *residual adapters* as base building for networks with a high-degree of parameter sharing. In this work, we propose some important improvements over this basic module. First, we show that a simple change, where the topology of the adapter is *parallel* rather than series, results in major improvements across the board, in terms of overall accuracy, applicability to existing off-the-shelf network, and transfer learning. Second, we investigate which parts of typical network require adaptation, and we show that often *both early and late layers* need to be adapted to obtain the best performance. Third, we experiment with different regularization strategies for the adapters such as dropout which proves highly beneficial when using a bigger pretrained network. Fourth, we introduce a cross-domain compression procedure for the adapters which allows to reduce significantly the numbers of adapters parameters. Most importantly, this compression contributes to multi-domain regularization resulting in improved overall performance thanks to information sharing among target datasets.

The rest of the paper is organized as follows. Section 2 discusses related work. Section 3 describes our neural network parametrization and how it applies to state-of-the-art neural network architectures. Section 4 demonstrates empirically the power of our approach on standard datasets, setting in particular the new state of the art on the Visual Decathlon benchmark, as well as demonstrating excellent transfer learning capabilities. Finally, section 5 summarizes our findings.

## 2. Related Work

Our work intersects with various lines of research in multi-task learning, learning without forgetting, domain adaptation, and other areas.

**Multi-task learning (MTL)** aims at learning multiple related tasks simultaneously by sharing information and computation among them. Early work [5] in this area focuses on deep neural network (DNN) models which share weights in the earlier layers and use specialized ones in the later layers. It is shown in [5] that sharing parameters during training helps exploiting regularities present across tasks and improving the performance by constraining the learned representation. However this setting requires to manually design the network and decide which layers should be shared across multiple tasks. This paradigm is applied to various learning problems from natural language processing [6] and automated drug discovery [7] to speech recognition [12]. In computer vision, deep MTL models are applied to object tracking [34], facial-landmark detection [35], object and part detection [2], object detection and instance segmentation [10], a collection of low-level and high-level vision tasks [14]. Differently from our work, this line of research focuses on learning a diverse set of tasks in the same visual domain.

**Multi-domain learning.** Our method is most related to recent works [3, 23, 25] which aim at learning a single network to perform image classification tasks in a diverse set of domains. The main focus is to learn a single network that can represent compactly all the domains with minimal number of task specific parameters. To do so, Bilen and Vedaldi [3] propose to model different domains in a sin-gle neural network by sharing all core model parameters except parameters in batch and instance normalization layers. Rebuffi *et al*. [23] extend [3] and propose a new parameterization of the standard residual network architecture that enables a high degree of parameter sharing between domains with a small increase ($< 10\%$) in the model parameters. The authors of [25] propose a parameter-efficient architecture that enables learning new domains sequentially without forgetting. We build our method on [23, 25] and significantly improve over them in terms of accuracy and compression ratio by introducing a novel and more compact adapter module, and a better regularization strategy.

**Parameterized MTL.** Another MTL approach [1, 32, 20] focuses on dynamically generating DNN weights given the task identity. Bertinetto *et al*. [1] propose a method to learn the parameters of a deep model from a single exemplar for one-shot classification. As a naive predicting of high dimensional weights is not feasible, the authors first obtain a low rank decomposition of filters and define the new network as a linear combination of the low-rank filters. Similarly, the authors of [32] propose a tensor factorization method that can realize automatic learning of end-to-end knowledge sharing in deep networks. Meyerson and Miikkulainen [20] propose a soft ordering approach, which dynamically computes to what extent each filter contributes to each tasks and thus how much is shared across different tasks. As a matter of fact, we also use a similar low rank decomposition technique to the one in [1]. However, the decomposition is used to design a more compact sharing across tasks.

**Domain adaptation.** There is a rich body of work in domain adaptation including the ones in deep learning such as [9, 31] that minimizes the domain discrepancy. The authors of [19] propose a deep network architecture that can jointly learn adaptive classifiers and transferable features from the source to target domain by modeling source classifier as sum of target classifier and a residual function. Bousmalis *et al*. [4] consider an explicit parameterization of domain-generic and domain-specific that learns to extract image representations from the partitioned subspaces. Li *et al*. [17] propose a meta-learning method that trains any given model to be more robust to domain shift. Our method differs to this group of work in two important aspects: First, in addition to domain change (*e.g*. DSLR *vs*. webcam), each domain contains a unique set of outputs (*i.e*. object categories) in our case. Second, domain adaptation typically aims to maximise performance on the target domain regardless of potential forgetting.

**Life-long learning.** Another important research direction in MTL is sequential learning of multiple tasks [21, 30]. While the key idea is to exploit the knowledge from the previous tasks, learning sequentially typically suffers from forgetting the previous tasks, a phenomenon referred as "catas-trophic forgetting" in [8]. Recent work [29, 26] address this problem by freezing the network parameters for the old tasks and only updating the parameters of the new task which leads to a linear growth in the number of total parameters with the number of tasks. Another approach is to preserve the previous knowledge by retaining the response of the original network on the new task [18, 24]. The problem is also addressed by keeping the network parameters [13] and features [22] of the new task close to the original ones. Our method can also be related to both [26, 13], as it retains the knowledge of previous tasks perfectly, while adding a small number of extra parameters for the new tasks.

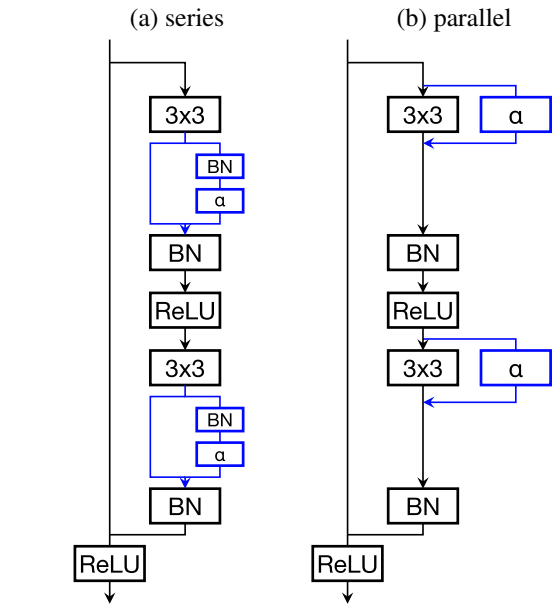

Figure 2: Series *vs* parallel residual adapters. (a) typical module of a residual network inclusive of batch normalization layers and residual adapters (in blue). (b) the same configuration, but with parallel adapters instead, resulting in a simpler network.

## 3. Method

This section describes different ways of constructing a parametric family of neural networks that can tackle multiple domains while sharing the vast majority of their parameters. Section 3.1 introduces a number of *adapter modules*. These modules attach to a standard deep neural network architecture such as ResNet [11] to steer it to different problems by means of a small number of adaptation parameters. Section 3.2 discusses different ways in which residual adapters can be injected in a standard neural network, section 3.3 how they can be regularized, and section 3.4 how the parameters can be further compressed.

## 3.1. Adapter modules

We begin by reviewing the recent adapter modules of [23] (section 3.1.1). We then discuss a number of alternative designs that, as shown empirically in section 4, perform significantly better (3.1.2). These modules are illustrated in fig. 2.

### 3.1.1 Series residual adapters

The residual adapter modules introduced by [23] consist of a $1 \times 1$ filter bank in parallel with a skip connection (fig. 2.a):

$$\mathbf{y} = \rho(\mathbf{x}; \alpha) = \mathbf{x} + \mathrm{diag}_1(\alpha) * \mathbf{x}.$$

If the input tensor has shape $\mathbf{x} \in \mathbb{R}^{H \times W \times C}$, then $\alpha \in \mathbb{R}^{C \times C}$ has $O(C^2)$ parameters. Here we use the operator $\mathrm{diag}_L(A) \in \mathbb{R}^{L \times L \times C \times D}$ to reshape a matrix $A \in \mathbb{R}^{C \times D}$ in a bank of "diagonal" filters:

$$[\mathrm{diag}_L(A)]_{vucd} = \begin{cases} A_{dc}, & v = u = (L-1)/2 + 1, \\ 0, & \text{otherwise.} \end{cases}$$

This operator transforms the matrix $A$ into a $1 \times 1$ filter bank embedded as the central element of a larger $L \times L$ filter bank by appending zeros around it ($L$ is assumed to be odd).

An advantage of this relatively cumbersome notation is that we can rewrite the module as a single filter:

$$\rho(\mathbf{x}; \alpha) = \mathrm{diag}_1(I + \alpha) * \mathbf{x}$$

The rationale for the additive parameterization is that the identity function is recovered if $\alpha = 0$. This is the case when a strong regularizer is applied on $\alpha$ during learning, shrinking the weights towards zero. In turn, this allow to easily control the adaptation strength, and thus generalization.

Residual adapters are installed in series with standard filter banks $\mathbf{f} \in \mathbb{R}^{L \times L \times C \times C}$ in the neural network. So for example a typical sequence is

$$\mathbf{z} = \rho(\mathbf{f} * \mathbf{x}; \alpha) = (\mathrm{diag}_1(I + \alpha) * \mathbf{f}) * \mathbf{x}.$$

This can also be interpreted as a low-rank decomposition of a filter bank $\mathbf{g}$, using $\mathbf{f}$ as a basis:

$$\rho(\mathbf{f} * \mathbf{x}; \alpha) = \mathbf{g} * \mathbf{x}, \quad [\mathbf{g}]_{vucd} = \sum_d (1 + \alpha_{dc})[\mathbf{f}]_{vucd}.$$

This also means that the adapters can be "fused" with the convolutional layer $\mathbf{f}$ by computing $\mathbf{g}$ explicitly, with no added evaluation cost at test time. However, this operation is difficult to undo, preventing from retargeting the network to another problem, which may be inappropriate in certain applications.

**Size of the adapters.** In this configuration, the adapter parameters are a fraction $C^2/L^2C^2 = 1/L^2$ of the filter bank parameters. For example, for a $3 \times 3$ filter bank, $L = 3$ and the adapters are 9 times smaller.

**Relationship to batch normalization.** For learning, it is customary to inject batch normalization (BN) layers in architectures, especially of the very deep variety such as ResNet. Figure 2.(a) illustrates a complete residual module, inclusive of BN, ReLU, convolution, and adapter layers for the series configuration.

### 3.1.2 Parallel residual adapters

While in the previous section adapters are installed in series with existing filter banks $\mathbf{f}$, we propose here an alternative configuration in which adapters are connected in *parallel* instead (fig. 2.b):

$$\mathbf{y} = \mathbf{f} * \mathbf{x} + \mathrm{diag}_1(\alpha) * \mathbf{x} = (\mathbf{f} + \mathrm{diag}_L(\alpha)) * \mathbf{x}.$$

Parallel adapters can also be interpreted as a low-dimensional parametrization of a filter bank $\mathbf{g}$:

$$\rho(\mathbf{f} * \mathbf{x}; \alpha) = \mathbf{g} * \mathbf{x},$$

$$[\mathbf{g}]_{vucd} = [\mathbf{f}]_{vucd} + \begin{cases} \alpha_{dc}, & v = u = (L-1)/2 + 1, \\ 0, & \text{otherwise.} \end{cases}$$

However, differently from series ones, in this case the decomposition is *affine*. The parameters $\mathbf{f}$ can be thought as a universal filter bank which is adjusted additively by modifying the "diagonal" elements of the filters based on $\alpha$.

Like for the series residual adapters, at test time it is possible to "fuse" the adapters $\alpha$ and filters $\mathbf{f}$ by computing $\mathbf{g}$ explicitly. Differently from that case, however, this additive change can be easily undone to allow to retarget the network to a new task.

**Size of the adapters.** If $\mathbf{f} \in \mathbb{R}^{L \times L \times C \times C}$, then $\alpha \in \mathbb{R}^{C \times C}$ has the same dimensions as before, so parallel and series adapters have the same number of parameters. It also benefits from the same shrinking to identity property, as setting $\alpha = 0$ recovers $\mathbf{f}$.

**Relationship to batch normalization.** Just as for series adapters, injection in a neural network such as ResNet [11] requires to clarify the relationship between the adapters and other layers such as BN. This is illustrated in fig. 2.(b). Note that the parallel configuration is *significantly simpler*. For example, compared to the parallel adapters, it saves one BN layer per application.

**Further discussion.** For both residual and parallel adapters, the filters $\mathbf{g}$ are points in a certain low-dimensional affine subspace parameterized by $\alpha$. However, for residual adapters the affine subspace is linear (passes through the origin) and its orientation is variable. For parallel adapters the subspace is affine and the orientation is fixed (given by coordinate axis along the "diagonal").

## 3.2. Network architecture

Having chosen a type of adapter modules, the next question is how they can be best applied to a deep neural network. Adapters may be applied throughout its depth, or more adaptation may be required at the shallower, intermediate, or deeper layers.

To explore these design strategies, we consider as baseline model ResNet [11] in the 26-layer configuration (suitable for medium-sized images). This network (section 3.2) is formed of 3 macro-blocks of convolutional layers, each outputting 64, 128 and 256 feature channels. Each macro-block contains 4 residual blocks each, each of which consists of two convolutional layers using $3 \times 3$ filters and a skip connection. The resolution of the data is halved from a macro-block to the next using average pooling. Note that, compared to other architecture such as AlexNet [16] and VGG16 [27], ResNet has a minimal fully-connected layer, meaning that abstraction is likely to increase more uniformly throughout the convolutional part of the network.

In order to experiment with different placements for the adapters, ResNet is broken down into three trunks: early, mid, and late, corresponding to the three macro blocks. Empirically (section 4), we apply the adapters to each stage individually, or to the three stages together. We also experiment with distributing adapters throughout the depth of the model, but skipping one every two, by adapting only the second convolutional layer in each residual block.

Note that the adapter dimensionality is determined by the number of channels in different layers of the architectures. Adapters applied to deeper layers are therefore bigger because the number of feature channels increases with depth. In section 3.4 we show how adapters can be further compressed.

## 3.3. Regularization: shrinkage *vs* dropout

One advantage of residual adapters is that they revert to the original neural network when $\alpha$ is zero. This is true for series adapter (as noted before) as well as for parallel adapters.

However, there are many alternative forms of regularization that apply to deep networks. For example, BN layers are noisy by construction, and are known to help regularize learning. Another well known method is dropout [28]. In the experiments, shrinkage is compared empirically against dropout, and the latter is shown to be necessary when using a bigger pretrained network. Note that, due to the additive nature of the adapter, dropout in this case is akin to injecting additive noise to the output of the network filters.

## 3.4. Cross-domain adapter compression

The size of a residual adapter is determined by the number of feature channels of the convolutional layer it is applied to. For deep layers in a neural network, where the number of channels $C$ can be quite large, the number $C^2$ of adapter parameters can still be non-negligible.

In order to address this issue, we propose to further compress the adapters. A simple approach is to consider a *low rank decomposition* $\alpha = \beta\gamma^\top$ of the adapter matrix $\alpha \in \mathbb{R}^{C \times C}$, where $\beta, \gamma \in \mathbb{R}^{C \times K}$ and $K \ll C$. Such a decomposition can be obtained efficiently using the SVD to minimize the reconstruction residual $\|\alpha - \beta\gamma^\top\|_F$. After replacing $\alpha$ with $\beta, \gamma$, the latter are fine-tuned again on the target task to improve performance further. This scheme uses a fraction $2KC/C^2 = 2K/C$ of the parameters.

Better compression can be obtained by decomposing the adapters jointly for all domains. In order to do so, let $\alpha_1, \ldots, \alpha_T \in \mathbb{R}^{C \times C}$ be domain-specific adapters for $T$ tasks. After stacking these matrices, computing the SVD decomposition of the result, and retaining only the top $K$ singular values, one gets:

$$\begin{bmatrix} \alpha_1 \ldots \alpha_T \end{bmatrix} = U\Sigma V = \begin{bmatrix} U \\ \vdots \end{bmatrix} \begin{bmatrix} \bar{\Sigma} & \\ & \ddots \end{bmatrix} \begin{bmatrix} \bar{V}_1^\top & & \bar{V}_T^\top \\ \vdots & | & \vdots \end{bmatrix}$$

where $U, \bar{V}_t \in \mathbb{R}^{C \times K}$, $U^\top U = \sum_t \bar{V}_t^\top \bar{V}_t = I \in \mathbb{R}^{K \times K}$ and $\bar{\Sigma} \in \mathcal{K} \times \mathcal{K}$ is diagonal. Setting $\beta = U\bar{\Sigma}$ and $\gamma_t = \bar{V}_t^\top$, we obtain the approximation:

$$\forall t = 1, \ldots, T: \quad \alpha_t \approx \beta\gamma_t^\top \qquad (1)$$

where $\beta, \gamma_t \in \mathbb{R}^{C \times K}$. In this case, $\beta$ is shared between domains acting as a common metric and only the factors $\gamma_t$ are fine-tuned to simplify optimization.

The total number of parameters in $(\beta, \gamma_1, \ldots, \gamma_T)$ over the parameters in $(\alpha_1, \ldots, \alpha_T)$ for a large number of tasks $T$ is given by

$$\frac{TCK + CK}{TC^2} \to \frac{K}{C}.$$

In practice, we show that good results can be obtained by setting $K = C/2$, therefore with a $2\times$ reduction in the adapter parameters. Joint compression also allows target tasks to communicate and further share parameters ($\beta$), in contrast with [23, 25] where adapters are independent. We show that this results in a multi-task regularizer which allows each domain to further benefit from the knowledge of the others.

Finally, note that the parallel adapters can be seen as a parametrization of filters spanning a fixed coordinate subspace. Equation (1) provides a more efficient parametrization of the same subspace, resulting in a higher degree of parameter sharing.

## 4. Experiments

This section thoroughly assesses the proposed designs, including the topology and position of the residual adapters

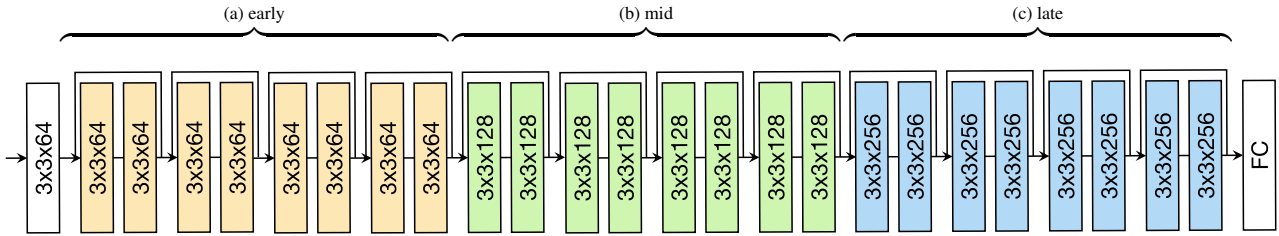

Figure 3: Adapter injection in ResNet-26. Following the scheme of fig. 2, adapters are added to each residual block (here given by a pair of convolutional blocks). We experiment with focusing adaptation on different segments of the network: early, mid, and late. adapters

and the regularization and compression strategies introduced in section 3. We evaluate these decisions quantitatively in multi-domain learning (section 4.1) and transfer learning scenarios (section 4.2). We share our code and models in https://github.com/srebuffi/residual_adapters.

### 4.1. Learning multiple domains

We first investigate the problem of learning multiple, visually-diverse domains using a parameterized neural network family. To this end, we use the recently-introduced Visual Decathlon benchmark [23]. This benchmark consists of 10 different well known datasets, from ImageNet, to OmniGlot (glyphs) and German Traffic Signs. In the benchmark, images are resized to a common resolution of roughly 72 pixels to accelerate evaluation. Furthermore, given the different nature and difficulty of the problems, results are reported both in terms of top-1 accuracy as well as using a "Decathlon score" that rebalances the different problems making them comparable [23] (see table 1).

Following [23], we first train the universal parameters **w** of the model using the ImageNet data with a 26 layer ResNet [11] via a stochastic gradient optimization with momentum and finally obtain 60.32% top-1 accuracy on 72 pixel resized validation set. As the first baseline, we finetune the pre-trained network for each dataset separately, denote it as "Finetuning" and report its performance in table 1. This standard procedure produces a strong baseline with competitive results, 76.9% mean accuracy and score 3096. However it requires ten times more parameter capacity than the base network, as it needs to train one network for each domain.

**Parallel vs Series.** Next, we compare different topologies for the adapter modules, series and parallel (see section 3.1). For both settings, we first freeze the weights of the pre-trained ImageNet model and learn only the adaptation parameters $\alpha_{1,\cdots,K}$ for each domain. Compared to fine-tuning, adding class-specific adapters lead to a modest increase ($2\times$ vs $10\times$) in total number of parameters. Despite their compactness, both approaches outperform the fine-tuning baseline, achieving similar or better accuracy over all datasets. This indicates that substantial parameter sharing is possible. We also see that the parallel configuration outperforms the series one (by 1 point in average accuracy and 250 decathlon points).

The parallel configuration has the key advantage of being *plug-and-play* whereas the series configuration of [23] requires the *adapters to be included* when ResNet is pre-trained on ImageNet. Indeed, adding them a-posteriori decreases performance substantially (-1.73 point in accuracy on average over 4 datasets). In contrast, parallel adapters can be appended to any pre-trained network, which allows them to be used with off-the-shelf models.

**Location of residual adapters.** Here we study the optimal placement strategy for the residual adapters throughout the network. As shown in fig. 3, the network is composed of three macro blocks, early, mid and late. In the first experiment, we apply the parallel residual adapters to each macroblock, skip the other two and report the results in table 1 as "Parallel (early,mid,late)". We observe that it is crucial to use the adapters in all the macro blocks as these 3 partial models perform significantly worse than the full model. Still, the adapters are most beneficial in the last block which suggests that, as expected, filters become more specialized and domain specific towards the end of network. We also investigate how the adapters should be distributed within each residual block. In the default setting, the adapters are applied at each of the two convolutional layers (see fig. 2). We evaluate the performance when it is applied to only the second convolutional layer which reduces the number of domain specific parameters by half. We observe that this results in a consistent drop in classification accuracy, suggesting that adapting each convolutional layer is beneficial.

**Regularization.** One of the challenges of training a single network for multiple tasks is to find an optimal training setting that can work when tasks differ in their difficulty level and number of training images. For instance, we observe in the preliminary experiments that training the adapter modules on the domains with fewer images per class such as Aircraft, DTD, Flowers datasets lead to overfitting on the training set after only a few iterations. To prevent this, we apply a stronger regularization by increasing the weight decay during training time. In particular, we group the datasets in terms of size of their training set as in [23] and assign a different weight decay value for each dataset *i.e.* higher weight decay for smaller datasets (0.002 for Aircraft, DTD, Flowers, 0.0005 for Omniglot, Pedestrian and UCF101 and 0.0001 for CIFAR100, GTSRB and SVHN). This forces a stronger regularization for smaller datasets such that the resulting network has to stay close to the pretrained network. In addition to shrinkage, we also evaluate the effect of another popular regularization strategy, dropout [28]. In this experiment, we apply dropout just before the second parallel adapter in each residual block as done in the standard WideResNet [33]. Figure 4a shows classification accuracies for parallel adapters (with and without dropout) used with pre-trained ResNet models with varying filter widths. We see that dropout needs a wider pretrained network (2.5×) to be effective and that the effect is significant no matter what the size of the training set with a state-of-the-art 85% accuracy using the full training set or an impressive 73% accuracy using only 50 images per class. Thus, dropout enables a better use of the adapters for high capacity pretrained networks even when few images per class are available.

| Model | #par. | ImNet | Airc. | C100 | DPed | DTD | GTSR | Flwr | OGlt | SVHN | UCF | mean | $S$ |
|---|---|---|---|---|---|---|---|---|---|---|---|---|---|
| # images | | 1.3m | 7k | 50k | 30k | 4k | 40k | 2k | 26k | 70k | 9k | | |
| Finetuning | 10× | 60.32 | 60.34 | 82.12 | 92.82 | 55.53 | 99.42 | 81.41 | 89.12 | 96.55 | 51.20 | 76.88 | 3096 |
| Series Res. adapt. | 2× | 60.32 | 61.87 | 81.22 | 93.88 | 57.13 | 99.27 | 81.67 | 89.62 | 96.57 | 50.12 | 77.17 | 3159 |
| Parallel Res. adapt. | 2× | 60.32 | 64.21 | 81.91 | 94.73 | 58.83 | 99.38 | 84.68 | 89.21 | 96.54 | 50.94 | 78.07 | 3412 |
| Parallel (early) | 2× | 60.32 | 50.47 | 78.58 | 93.26 | 58.46 | 99.00 | 82.27 | 87.68 | 95.39 | 47.77 | 75.32 | 2610 |
| Parallel (mid) | 2× | 60.32 | 57.88 | 79.25 | 94.24 | 56.65 | 98.85 | 83.43 | 88.47 | 95.96 | 48.98 | 76.40 | 2852 |
| Parallel (late) | 2× | 60.32 | 61.06 | 80.58 | 94.02 | 57.87 | 99.19 | 84.68 | 89.06 | 96.30 | 50.94 | 77.40 | 3159 |
| Parallel (half) | 1.5× | 60.32 | 61.15 | 81.24 | 94.36 | 58.40 | 98.85 | 84.76 | 88.69 | 96.19 | 49.99 | 77.40 | 3061 |
| Parallel SVD | 1.5× | 60.32 | 66.04 | 81.86 | 94.23 | 57.82 | 99.24 | 85.74 | 89.25 | 96.62 | 52.50 | 78.36 | 3398 |
| Rebuffi *et al.* [23] | 2× | 59.23 | 63.73 | 81.31 | 93.30 | 57.02 | 97.47 | 83.43 | 89.82 | 96.17 | 50.28 | 77.17 | 2643 |
| Rosenfeld & Tsotsos [25] | 2× | 57.74 | 64.11 | 80.07 | 91.29 | 56.54 | 98.46 | 86.05 | 89.67 | 96.77 | 49.38 | 77.01 | 2851 |

Table 1: Reports the (top-1) classification accuracy (%) and decathlon overall score ($S$) of different models on the decathlon tasks [23]. The model size ("#par") is the number of parameters w.r.t. the vanilla network pretrained on ImageNet. Our best models use the parallel adapters and SVD, indicated as "Parallel SVD".

**Adapter compression.** The size of each residual adapter is dictated by the number of filters in its corresponding convolutional layer. In most of the modern deep network architectures such as AlexNet [15], ResNet [11], the number of convolutional filters is designed to double after each block. This leads to significant increase in the adapter size at the later layers. While this is found to be beneficial for a generic network design, we speculate that the dimensionality of required residual modules can be reduced without any drop in classification performance, as some filter combinations can be useful for more than one domain. Thus, we assume that weights of adapter modules $\alpha$ for different domains are not linearly independent. To test our reason-ing, we first take the pre-trained ResNet model and freeze all the weights and only learn domain specific parameters $\alpha$. As described in section 3.4, we stack these weights, apply the SVD and only retain half of the original dimensionality that yields a further 50% reduction in parameters. Finally, we freeze $\beta$ weights and fine-tune $\gamma$ for each domain. We show in table 1 that this approach preserves the performance of the default parallel residual modules while having lower number of parameters (twice as less adapters parameters). As expected, the cross-domain compression acts as a multi-task regularizer and thus prevents from overfitting on small datasets. For example, we can point out the significant effect on Aircraft, VGG-Flowers or UCF 101 with respectively an improvement of 1.8, 1.1 and 1.5 accuracy points with less trainable parameters. For bigger datasets, the performances are preserved while reducing the number of parameters.

**Comparison to the state-of-the-art.** We also compare our method to the recent work [23, 25] that report results on the Decathlon dataset and show in table 1 that our method significantly outperforms both. Our approach is directly comparable to [23] as the same base network is used; in particular, "Series Res. adapt." in table 1 is our re-implementation of [23], which outperforms the original in terms of mean accuracy and Decathlon score (3159 vs 2643). Our final result (SVD) achieves a boost of 1.2% in classification accuracy and approximately 750 Decathlon points while employing only half of the additional parameters used in [23]. Similarly, we obtain a remarkable improvement over [25] (1.4% in classification accuracy and approximately 550 Decathlon points) with a significantly more compact architecture.

### 4.2. Transfer learning results

A desired property for a multiple domain learning method is the ability to learn a previously unseen domain

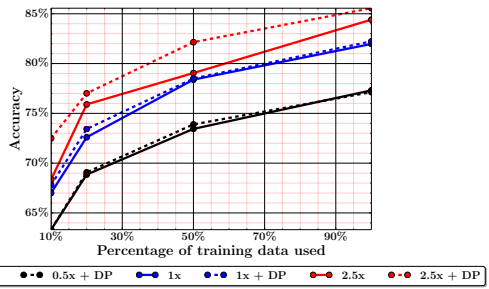
(a) Different pretrained network widths w/ and w/o Dropout

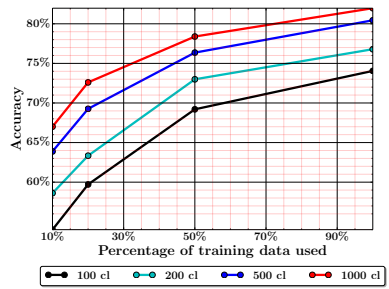
(b) Influence of number of pretrained classes

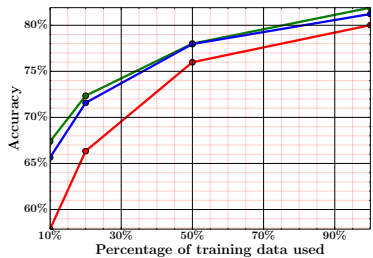
(c) CIFAR 100

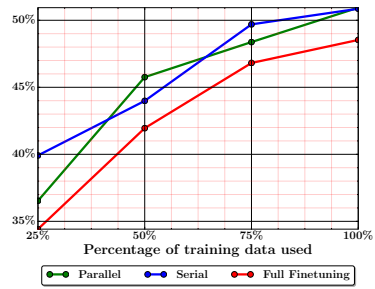
(d) UCF 101

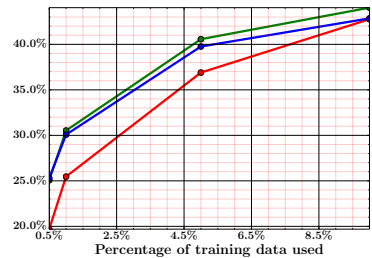
(e) MIT Places

Figure 4: (a,b) analyze on CIFAR100 the influence of pretrained network settings when combined with parallel adapters. (c,d,e) compare the performances of the different methods on 3 datasets in the Transfer Learning setting.

especially when training data is limited. To assess this quantitatively, we take the pretrained ResNet model on the ImageNet and finetune by using the residual adapters on three datasets, UCF-101 ("small"), CIFAR-100 ("medium") and MIT Places 205 ("large") with more than 2 million images, all resized to 72 pixels. We train our method with varying the percent of training data and report the results in Figure 4. Both the parallel and series configurations clearly outperform finetuning, not only when there is fewer data available but also for the full size of CIFAR-100 and UCF101. Finetuning only outperforms our method when it is trained on the full training set of the MIT Places and obtains $51.13\%$ compared to our $47.2\%$ validation accuracy. As our method only updates the adapter parameters, finetuning can exploit the high capacity of the network. Hence, the parallel adapters compare very favorably to standard fine-tuning except for extremely large datasets. Series adapters are similar, but with the key difference that the parallel configuration can be applied to an off-the-shelf model a-posteriori. In short, parallel adapters are a simple strategy that can replace and outperform standard fine-tuning in almost every way.

### 4.3. Influence of the pre-training network

We discussed previously that dropout allows an efficient use of wider networks with the parallel adapters and fig. 4a shows that increasing the pretrained network size (from $0.5\times$ to $2.5\times$) helps even when amount of training data is limited. Here we also study how the pretrained network affects the performances of transfer learning when it is trained on a training set sampled from a smaller number of categories. We observe in fig. 4b that the classification accuracies on the target task decrease steadily if we pretrain the same network with less ImageNet classes. Thus a good network for transfer learning with adapters should be saturated with as many classes as possible during pretraining.

## 5. Conclusion

In this paper we have shown that it is possible to build universal parametric families of networks that can share parameters very efficiently among multiple domains. We have proposed and evaluated several design strategies for the design of such architectures. The best results were obtained by using parallel residual adapter modules distributed throughout a neural network architecture and further jointly rank-compressed. The resulting network families are very compact, resulting in substantial savings in terms of model storage, exchange, update, and transmission. They also significantly outperform recent alternatives in benchmarks such as Visual Decathlon. We have also showed that parallel adapter can replace traditional fine-tuning techniques, achieving far superior performance that those in almost all cases with no additional constraint or limitation.

**Acknowledgments:** This work acknowledges the support of Mathworks/DTA DFR02620, EPSRC SeeBiByte and ERC 677195-IDIU.